# Advancing Recycling Efficiency: A Comparative Analysis of Deep Learning Models in Waste Classification

Zhanshan Qiao

School of Computer Science & Technology, Beijing Institute of Technology, Beijing,China

1120211577@bit.edu.cn

**Abstract**.With the ongoing increase in the worldwide population and escalating consumption habits,there's a surge in the amount of waste produced.The situation poses considerable challenges for waste management and the optimization of recycling operations.The research tackles the pressing issue of waste classification for recycling by analyzing various deep learning models,including Convolutional Neural Network(CNN),AlexNet,ResNet,ResNet50 plus Support Vector Machine(SVM),and transformers,across a wide array of waste categories.The research meticulously compares these models on several targets like parameters settings,category accuracy,total accuracy and model parameters to establish a uniform evaluation criterion.This research presents a novel method that incorporates SVM with deep learning frameworks,particularly ResNet50.The results indicate the method significantly boosts accuracy in complex waste categories.Moreover,the transformer model outshines others in average accuracy,showcasing its aptitude for intricate classification tasks.To improve performance in poorly performing categories,the research advocates for enlarging the dataset,employing data augmentation,and leveraging sophisticated models such as transformers,along with refining training methodologies.The research paves the way for future advancements in multi-category waste recycling and underscores the pivotal role of deep learning in promoting environmental sustainability.

## 1. Introduction

As the global population continues to rise and consumption patterns intensify,the volume of waste generated is increasing.It brings significant challenges in waste management and optimizing recycling processes.According to the Global Waste Management Outlook 2024 from the United Nations Environment Programme(UNEP),the global municipal solid waste generation is predicted to grow from 2.1 billion tonnes in 2023 to 3.8 billion tonnes by 2050.In view of the huge amount of waste,the automation of waste sorting has emerged as an essential component of the recycling workflow.The implementation of smart waste sorting technologies can not only reduce the burden on manual labor but also significantly enhance the accuracy of recycling,leading to higher recovery rates of valuable materials.Furthermore,it aid in efficient resource utilization,cutting down the overall environmental footprint and contributing towards a circular economy.

    One of the most viable and forward-looking methods in waste classification is through the use of computer vision technology.It have already seen widespread adoption across various sectors,with

applications ranging from facial recognition and fingerprint analysis,to the identification of cancerous cells and the innovation of autonomous vehicles [1]. Deep learning is notably applied in image classification,bypassing manual feature design and using deep neural networks to directly extract features and classify images.This method promises a more efficient and sophisticated waste segregation,improving recycling processes.Leveraging the advancements of computer vision technologies powered by deep learning algorithms,the potential to transform and streamline the trash sorting process is tremendous [2]. For example,a reliable and robust deep learning model has surpassed several advanced models by achieving an impressive accuracy rate of 95.01 percent on the TrashNet dataset [3]. It has also garnered high F1-scores across all six waste categories,scoring 96.18% for glass,97.24% for cardboard,95.73% for paper,94% for metal,93.67% for plastic,and 88.55% for litter[3].Moreover,according to a novel framework for trash classification using deep transfer learning,the improvement of ResNext model named DNN-TC was developed which is an attempt to improve the predictive performance [4]. DNN-TC achieved an accuracy of 98% and surpassed the performance of current state-of-the-art methods in trash classificationon on the VN-trash dataset,which comprises 5904 images categorized into three classes:Organic,Inorganic, and Medical wastes from Vietnam [4].

Although these models and algorithms achieve high accuracy on their respective datasets,the categories of waste in these datasets are generally limited.TrashNet only includes six categories and VN-trash dataset only includes three categories. Up to now,Studies involving a larger number of categories are relatively rare.Hence,this research utilizes a self-created dataset comprising 16 categories. By using models of CNN, AlexNet, ResNet, ResNet50 plus SVM,Transformer on the dataset, the research investigates which of these foundational approaches yields better performance in multi-category waste classification. At the same time,the research compares the performance of different models across varying sample sizes based on the accuracy of different waste categories.The research is intended to provide a foundational basis for future research on multi-category waste classification.

The contributions of this paper can be summarized as follows:

A significant contribution of the research is the innovative combination of ResNet50 with SVM.The hybrid model capitalizes on the deep representational power of ResNet50 and the precise classification capabilities of SVM,demonstrating a notable improvement in accuracy for challenging waste categories.The extensive evaluation results offers a unique insight into the performance of each model across a wide range of waste categories,establishing a benchmark for future research in the domain.

Through comprehensive experiments,the research demonstrates the superior performance of the ResNet50+SVM model in accurately classifying challenging waste categories,highlighting the model's practical efficacy.The experiments provide empirical evidence of the potential of advanced models like transformers in complex classification tasks,contributing to understanding of effective deep learning applications in environmental sustainability efforts.

## 2. Related work

In recent years,the application of machine learning and deep learning techniques has significantly advanced the field of waste classification and recycling,leading to more efficient and automated processes.In the realm of waste classification,several studies have adopted and compared various machine learning and deep learning approaches to enhance accuracy and efficiency.

Several researchers have focused on leveraging traditional machine learning algorithms for waste classification.In 2020,Adita Putri Puspaningrum and colleagues developed a waste image classification technique,using SVM for classification and combining Scale Invariant Feature Transform (SIFT) with Principal Component Analysis(PCA) for efficient feature extraction.This approach leverages SIFT to extract features and PCA to reduce feature dimensionality [5].

Moreover, in 2023, Deepika Kamboj et al. undertook a comprehensive comparison of several algorithms for waste classification,including SVM,Random Forest,Naïve Bayes Classifier,Decision

Tree,and KNN [6]. They evaluated each algorithm's suitability for waste classification based on accuracy,providing a broad view of their applicability in this context. Furthermore,in 2023,Ashish Pandey et al. delved into using CNN for waste classification,contrasting it with SVM across various waste types.Despite SVM's higher accuracy in some scenarios,the research underscored challenges in CNN's hyperparameter optimization [7]. However,they posited that CNN could potentially exceed SVM with these challenges addressed. Their work also integrated VGG16 and FastNet-34 architectures into a model tested on a Kaggle dataset of 22,564 images,showcasing improved classification accuracy [7].

Also in 2023, Jayati Bhadra and Aaran Lawrence DLima explored the efficacy of deep learning models,particularly CNN,VGG16,and ResNet50, pre-trained on the ImageNet dataset for internet-sourced waste image classification [8]. They highlighted ResNet50's superior performance,attributing it to its depth and pre-training,and suggested directions for future research. Additionally, in 2022, Aidan Kurz et al. introduced a novel algorithm combining Vision Transformers(ViT) and CNN into a Multi-Head block for parallel processing,aimed at solid waste object classification. This approach achieved a high accuracy with significantly reduced parameters compared to existing methods,emphasizing its potential for low-power applications and ease of deployment [9].

These contributions reveal a vibrant exploration of both machine and deep learning techniques in waste classification.While traditional machine learning algorithms offer a solid foundation for such tasks,the advent of deep learning,particularly with CNN and innovative architectures like ViT,opens new avenues for improving classification accuracy and operational efficiency.The ongoing challenge of optimizing model parameters,especially in deep learning applications,remains a critical area for future investigations to unlock the full potential of these technologies in waste management.

## 3. Method

The research delves into image classification aimed at advancing waste sorting and recycling,utilizing an innovative dataset that spans 16 distinct waste categories.Setting it apart from many existing public datasets that feature a limited diversity,this specially curate dataset offers an extensive array of waste types,crucial for practical recycling endeavors.

### 3.1. Data Collection

The dataset consists of images captured in varied environments to guarantee broad category representation.It covers a wide scope of waste from household trash to commercial and industrial debris,providing a rich variety for the thorough training and evaluation of algorithms.The dataset includes categories such as:old books,cans,iron wires,old shoes,old bags,ceramics,fruit peels,spent batteries,old foam, leftover food, paperboard, tissues, plastic bags, plastic bottles, glass bottles, shattered glass.

A meticulous selection process filters through this vast array of images.This crucial step eliminates images of low quality,under poor lighting,or those that are obscured,ensuring the inclusion of only crisp,representative samples in the dataset. The number of chosen images for each category totals 5960.The specific data distribution is shown in Figure 1.

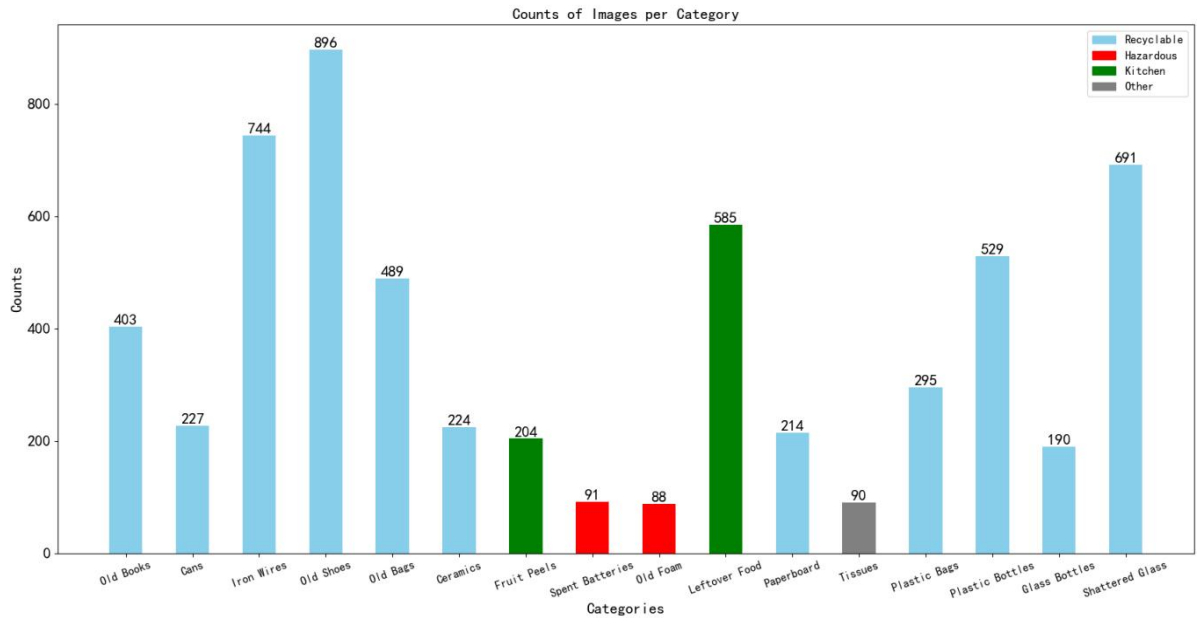

**Figure 1.** Counts of Image per Category (Picture credit : Original)

*3.2. Model Introduction*

*3.2.1. CNN* CNN:This proven and reliable architecture excels at recognizing spatial hierarchies within images (Figure 2).
 Transformer:This model is gaining recognition for its ability to effectively discern global relationships within data.

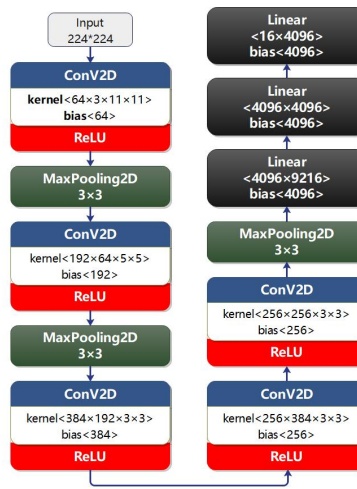

**Figure 2.** The structure of CNN [8]

 The model starts with a convolutional layer (Conv2d) that has a kernel size of 11x11, stride of 4 and padding of 2. This layer extracts initial features from the input image.Each convolutional operation is followed by a ReLU activation function which introduces non-linearity,enabling the network to capture complex features.To downs ample the feature maps,max pooling layers(MaxPool2d) are employed after certain convolutional layers. This reduces the spatial size of the representations,making the network more efficient and invariant to small translations of the input.
 Following the convolutional stack,an adaptive average pooling layer reduces the spatial dimensions of the feature maps to a predetermined size (6x6). This ensures a consistent input size to the classifier.

The classifier begins with a dropout layer which randomly zeroes some of the elements of the input tensor with probability equal to the dropout rate during training. This helps to prevent overfitting.A series of fully connected layers (Linear) map the pooled features to the final classification scores for each class. The first fully connected layer reshapes the flat tensor to a higher-dimensional space (4096 units), and the final layer reduces it down to the number of classes (specified by num_classes). The forward method of the CNN class defines the computation performed at every call of the model. It executes the sequence of operations outlined in the feature,avgpool,and classifier layers,effectively taking an input image and processing it through the network to produce class scores,from which the predicted class can be derived.

*3.2.2. ResNet50 + SVM* ResNet50 stands as a remarkable iteration in the evolution of neural network architectures,chiefly utilized as a backbone for numerous other neural networks due to its deep structure and residual learning framework [10]. The essence of ResNet50's design is its distinctive ability to learn from residual mappings alongside the layer inputs,which crucially mitigates the vanishing gradient issue prevalent in deep neural networks. Its depth fosters a broadened generalization capability,thereby enhancing performance across various complex datasets.

On the other side of the combination is the SVM,a model renowned for its classification prowess. SVM's core principle revolves around the optimization of margins:it meticulously constructs the optimal hyperplane to segregate data points,handling non-linear divisions adeptly.This margin optimization empowers the SVM with high discriminative competence,granting it notable precision in its classifications.The success of SVM can be attributed to the ability to harness extreme cases within the dataset,which lie close to the decision boundary [11].

Incorporating ResNet50 into a synergistic framework with SVM harnesses the advantageous traits of both models [11]. ResNet50's capability in deciphering layered data features yields an enriched initial data representation.This detailed feature set is then adeptly handled by the SVM,which sharpens the classification boundaries with its superior margin optimization methods.The rich feature-set extracted by ResNet50 ensures that the SVM has a high-quality input for creating robust classification models.

In essence, this combination is designed to capture the best of both worlds:the deep, nuanced understanding of data that ResNet50 provides,and the focused,precise classification ability of SVMs.

*3.2.3. Transformer* The Figure 3 describes the ViT structure diagram.The Vision Transformer adapts the Transformer structure,originally developed for natural language processing(NLP) tasks,to handle image classification challenges [12]. It fundamentally views an image as a sequence of patches,akin to the way sentences are composed of words in NLP [12].

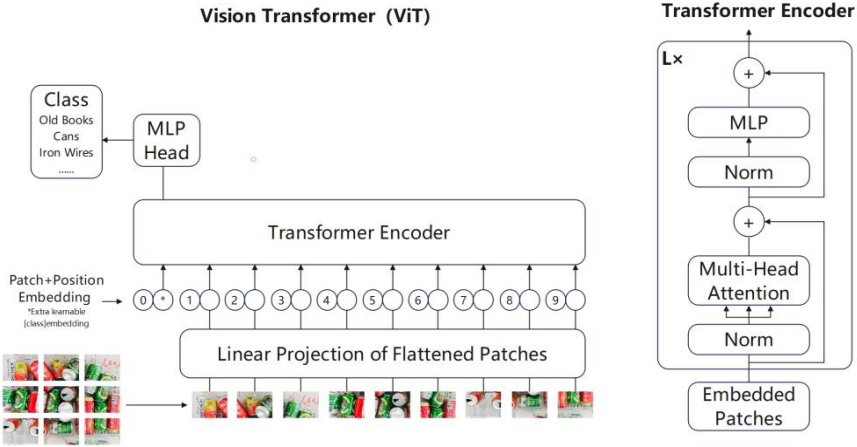

**Figure 3.** The structure of Vision Transformer(ViT) [12]

The remarkable achievements of transformers in NLP have inspired researchers to adapt this technology for computer vision[13].The Vision Transformer algorithm is a significant result of such efforts and has emerged as a landmark development in the CV domain[13]. Its capacity for global interaction has allowed it to outperform traditional CNN models[13]. The ViT model architecture is shown in Figure 3. Below are the essential elements of its design:

Image Tokenization:This process breaks down an image into uniform patches.These patches are then flattened and converted into a sequence of embeddings.Additionally,a unique "class token" embedding is incorporated into this sequence to symbolize the image as a whole.

Positional Encodings:To preserve the spatial ordering among patches,positional embeddings are integrated with the patch embeddings.This step is crucial for the model to recognize the arrangement and location of each patch within the image.

Transformer Encoder:This involves passing the embedding sequence through multiple Transformer encoder layers.Each layer employs a self-attention mechanism that enables the model to selectively concentrate on various segments of the image.A feed-forward neural network follows this.The architecture also employs residual connections and layer normalization consistently.

Classification Head:The class token's embedding,which now contains aggregated information across the image,is funneled through a linear layer to derive the ultimate classification result.

ViT's groundbreaking advancement is its utilization of the self-attention mechanism on image patches.This approach allows it to comprehend global interactions throughout the image efficiently and adaptively.

## 4. Result

### 4.1. Data set and experimental details

*4.1.1. Data set* The study explores image classification with the aim of enhancing waste sorting and recycling processes,utilizing a unique dataset that encompasses 16 different waste categories. Unlike many publicly available datasets, which often lack diversity,this carefully curated dataset provides a comprehensive range of waste types,essential for practical recycling applications.

The dataset comprises images taken in various environments to ensure a wide representation of waste categories,including household trash,commercial,and industrial refuse.It encompasses a diverse array of items such as old books,cans,iron wires,old shoes,and more,totaling 5960 selected images across all classes. Each category has a specific number of images ranging from 91 to 896,chosen through a rigorous selection process that discards low-quality images,ensuring the dataset consists of clear and representative samples ideal for in-depth algorithm training and assessment (Figure 4).

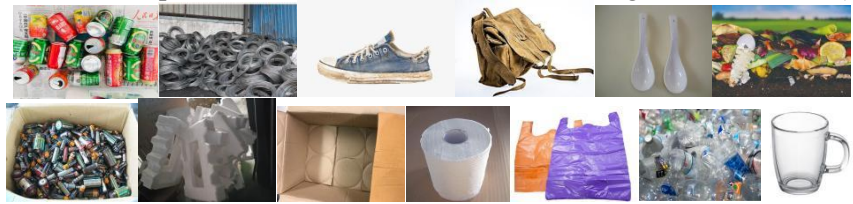

**Figure 4.** The samples of Dataset

*4.1.2. Experimental evaluation indicators* Accuracy serves as a critical metric for assessing the overall correctness of model predictions in classification tasks,reflecting the model's generalization ability to unseen data. Concurrently,the number of parameters influences the model's complexity and computational requirements,with an excess potentially leading to overfitting and a deficiency compromising the ability to capture complex patterns. This paper compares the performance and efficiency of different model approaches by examining these two key indicators: accuracy and the number of parameters.

*4.1.3. Parameter settings* Normalizing images to 224x224 matches pre-trained model requirements,optimizing visual feature preservation and computational efficiency. A low learning rate of 0.0001 avoids early local minima,enhancing stable training. The batch size of 32 balances training speed with computational demands,ensuring smooth progress. Setting epochs at 30 provides adequate learning time for pattern recognition,balancing efficiency with performance. A slight weight penalty prevents overfitting,crucial for complex models and large datasets. These parameters collectively ensure stable, efficient training and optimal performance across various model architectures (Table 1).

**Table 1** The parameter settings

| | Method/parameter settings | | | | |
|---|---|---|---|---|---|
| | CNN | AlexNet | ResNet | ResNet50 Plus SVM | ViT |
| Normalized | 224*224 | 224*224 | 224*224 | 224*224 | 224*224 |
| Learning rate | 0.0001 | 0.0001 | 0.0001 | 0.0001 | 0.0001 |
| Batch size | 32 | 32 | 32 | 32 | 32 |
| epochs | 30 | 30 | 30 | / | 30 |
| Weight Decay | 0.0001 | 0.0001 | 0.0001 | / | 0.0001 |

*4.2. Comparison of results*

*4.2.1. Performance comparison of each category* Table 2 shows the accuracy of different Categories from different methods.

**Table 2** The category accuracy

| | Method/Category accuracy(%) | | | | |
|---|---|---|---|---|---|
| | CNN | AlexNet | ResNet | ResNet50 Plus SVM | ViT |
| old books | 25% | 78% | 78% | 92% | 91.7% |
| cans | 15% | 54% | 80% | 75% | 88.4% |
| iron wires | 45% | 85% | 93% | 93% | 94.6% |
| old shoes | 51% | 86% | 90% | 95% | 95.2% |
| old bags | 53% | 82% | 84% | 94% | 98.6% |
| ceramics | 17% | 56% | 71% | 88% | 85.3% |
| fruit peels | 33% | 56% | 42% | 60% | 74.2% |
| spent batteries | 10% | 64% | 57% | 71% | 60.7% |
| old foam | 22% | 63% | 85% | 70% | 81.5% |
| leftover food | 56% | 88% | 94% | 89% | 85.8% |
| paperboard | 52% | 85% | 89% | 83% | 95.4% |
| tissues | 32% | 64% | 54% | 82% | 82.1% |
| plastic bags | 46% | 73% | 81% | 94% | 95.5% |
| plastic bottles | 55% | 73% | 84% | 85% | 89.9% |
| glass bottles | 70% | 84% | 91% | 98% | 96.5% |
| shattered glass | 70% | 92% | 93% | 93% | 98.1% |

- *Overall Analysis*

The dataset shows a significant class imbalance, ranging from 88 samples (old foam)to 896 (old shoes). When training a neural network, the presence of imbalanced classes results in unequal training frequencies across the different categories [14]. The imbalance in training leads to a significant disparity in the neural network's classification performance on test and validation datasets across various classes [14]. Consequently,categories that undergo more training iterations tend to yield better test outcomes,while those with fewer training instances tend to have inferior results.In general,traditional CNN method performs significantly worse than AlexNet, ResNet, ResNet50 plus SVM,and transformers across most categories. ResNet achieves better results than CNN and AlexNet in most categories,indicating that deeper network architectures can learn more complex features,thereby improving recognition accuracy. Combining the strengths of deep learning and traditional machine learning,ResNet50 plus SVM performs exceptionally well in certain categories,such as old books, iron wires, old shoes, old bags, plastic bags, glass bottles, and shattered glass. Transformers models achieve the best results in most categories,due to their powerful attention mechanism,which allows them to better capture global relationships and key features within images.

- *Category-Specific Analysis*

Categories with Poorer Performance(most methods below 80% accuracy):Old books,cans,ceramics,fruit peels,spent batteries,old foam,and tissues.The classification of these categories present the following challenges:high intra-class variance and high similarity to other categories.Given that classification models essentially function by matching templates,overlooking intra-class uncertainty or variability,particularly in datasets with imbalanced classes,can result in classification inaccuracies[15].For example,old books can vary significantly in appearance,making them harder to recognize.Fruit peels and leftover food may share similar colors and textures,making them difficult to distinguish.Relatively Insufficient samples in the dataset can lead to poor model generalization.

Categories with Better Performance(most methods above 80% accuracy):Iron wires,old shoes,old bags,leftover food,paperboard,plastic bags,plastic bottles,glass bottles,and shattered glass.The classification of these categories possess the following characteristics:low intra-class variance and Distinctive features.For example,plastic bottles tend to have uniform shapes and colors.Glass bottles have reflective properties,while iron wires have unique shapes.Relatively Sufficient samples in the dataset enables the models to learn more accurate features.

According to results of AlexNet,ResNet,ResNet50 plus SVM,and transformers,the examples can be roughly divided into four categories:High Accuracy&High Sample Size,High Accuracy & Low Sample Size,Low Accuracy & High Sample Size,Normal.

*4.2.2. Performance indicator analysis* Based on the table 3 provided,which shows the total accuracy and total number of parameters for various methods,we can analyze and compare the performance and complexity of each model:

**Table 3** The total accuracy and total parameters

| | \multicolumn{5}{c}{Method/Total accuracy and Total parameters} | | | | |
|---|---|---|---|---|---|
| | CNN | AlexNet | ResNet | ResNet50 plus SVM | ViT |
| Total accuracy | 51% | 79% | 84% | 89% | 92.7% |
| Total parameters | 5.7069M | 6.1101M | 2.3541M | 2.3508M | 8.5811M |

Basic CNNs often show lower accuracy due to their simple structure,while AlexNet improves upon this by adding depth and using advanced techniques like ReLU and dropout for better learning. ResNet outperforms both by efficiently using fewer parameters through residual connections,avoiding the vanishing gradient problem. Combining ResNet50 with SVM achieves even higher accuracy, effectively leveraging ResNet50's feature extraction with SVM's classification strength,and slightly reducing the overall parameter count by replacing the last layers with an SVM for optimized performance. Transformers has the highest total accuracy,which can be attributed to their ability to model long-range dependencies and focus on the most relevant parts of the image through the self-attention mechanism.However,it also have the most significant number of parameters,which implies that while they are powerful,they are also computationally heavy and may require more data and resources to train effectively. The higher total accuracy of AlexNet,ResNet,and Transformers compared to the basic CNN indicates that increased model capacity(more layers,parameters,and sophisticated architectures)can lead to better learning and generalization.ResNet's ability to achieve higher accuracy with fewer parameters compared to AlexNet and CNN suggests that how the parameters are used(residual connections in this case)can be more important than the sheer number of parameters.

In summary,the analysis reveals that deeper and more complex models,especially those with architectural improvements or hybrid strategies,tend to achieve higher accuracy.Transformers,despite their high parameter count,lead in terms of accuracy,which reflects the current trend in deep learning where attention-based models are setting new benchmarks for various tasks.

### 4.3. Visualization and analysis of results

Each result is composed of two plots. Left plot shows the overall accuracy of the model on the validation set as training progresses. The x-axis represents the training epochs,and the y-axis represents the accuracy.Right plot shows the loss of the model during training.The x-axis represents the training epochs,and the y-axis represents the loss.The blue line represents the training loss,and the orange line represents the validation loss.

Upon comparing the validation accuracy and loss curves of different neural network architectures in Figures 5 through 8,we observe the following: In Figure 5,the CNN model's validation accuracy rapidly increased initially and then slowed down,exhibiting mild fluctuations,while its validation loss began to rise after 20 cycles,indicating potential overfitting. In Figure 6,the AlexNet model's validation accuracy rose sharply within the first five cycles and then stabilized,with both training and validation losses showing a favorable trend of decline and stabilization. Figure 7 shows that the ResNet model's accuracy experienced significant fluctuations but overall trended upwards,with the validation loss demonstrating variability. In Figure 8,the transformer model's accuracy started relatively high,experienced initial fluctuations,and then reached a stable high level,with the validation loss showing slight fluctuation. Overall,although the CNN and ResNet models exhibited signs of overfitting and variability,the AlexNet and transformer models demonstrated more stable performance with higher accuracy and lower loss. These performance differences may be attributed to the distinct architectures,parameter settings,and training data used by each model.

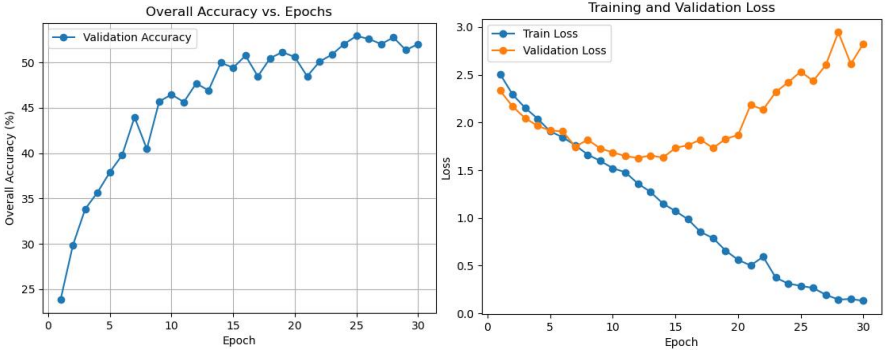

**Figure 5.** Validation accuracy and loss of CNN (Picture credit : Original)

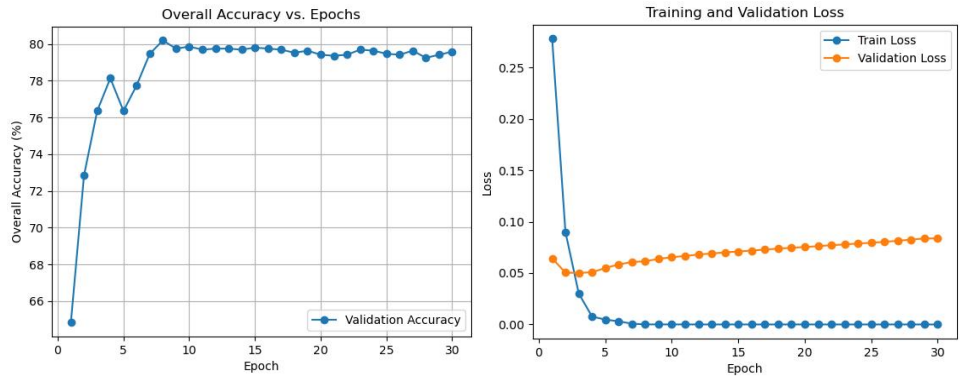

**Figure 6.** Validation accuracy and loss of AlexNet (Picture credit : Original)

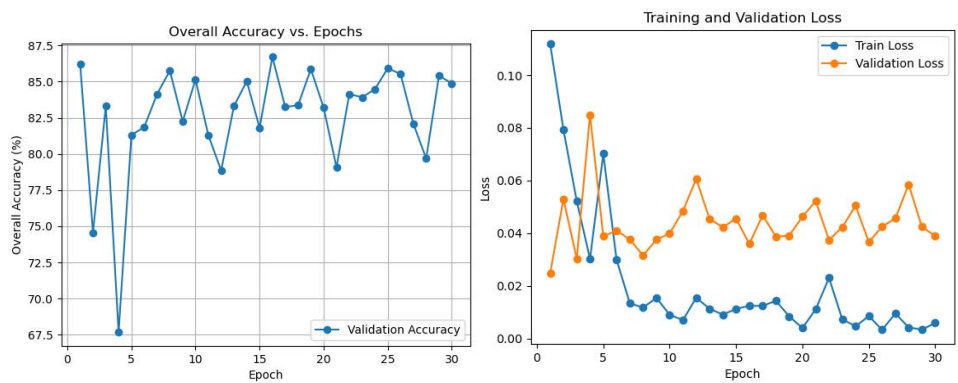

**Figure 7.** Validation accuracy and loss of ResNet (Picture credit : Original)

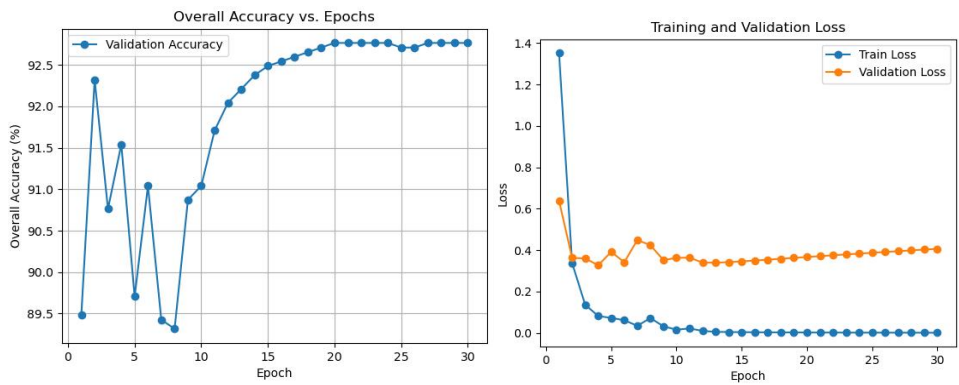

**Figure 8.** Validation accuracy and loss of Transformers (Picture credit : Original)

## 5. Conclusion

The research endeavors to address the critical issue of waste classification for recycling by conducting a comprehensive comparison of various deep learning models.To this end,an array of models including CNN,AlexNet,ResNet,ResNet50 plus SVM,and transformers were examined,utilizing a diversified dataset that represents a multitude of waste categories.Each model underwent scrutiny over several parameters,such as normalization,learning rate,batch size,epochs,and weight decay to ensure a consistent evaluation framework.

The experimental results revealed that models integrating SVM with deep learning structures like ResNet50 substantially improved category-specific accuracy,with the most notable increases observed in traditionally challenging classes such as 'old bags' and 'shattered glass'. Furthermore,the transformer model demonstrated the highest average accuracy,underscoring its potential for handling complex classification tasks.

The research concludes that for categories not performing optimally,several strategies can significantly enhance results.Accumulating a broader dataset,especially for classes demonstrating considerable variation,emerges as a crucial step.Employing data augmentation methods such as rotation and scaling is identified as effective in expanding dataset diversity.The utilization of more sophisticated models,like transformers,which excel in identifying complex patterns,alongside fine-tuning the training methodology and model parameters,is also highlighted as beneficial for improving classification outcomes.

Overall,the research sets the stage for further studies in multi-category waste recycling,identifying effective techniques and providing valuable insights to enhance automated sorting technologies.It is vital for achieving sustainable waste management.Future work could extend these benchmarks by adding more types of waste and testing these models in real-time scenarios. Such advancements will likely improve the models and affirm the importance of deep learning in supporting environmental sustainability efforts.